\documentclass[lettersize,journal]{IEEEtran}
\usepackage{amsmath,amsfonts}
\usepackage{algorithmic}
\usepackage{algorithm}
\usepackage{hyperref}
\usepackage{orcidlink}
\usepackage{array}
\usepackage[caption=false,font=normalsize,labelfont=sf,textfont=sf]{subfig}
\usepackage{textcomp}
\usepackage{color}
\usepackage{stfloats}
\usepackage{url}
\usepackage{verbatim}
\usepackage{graphicx}
\usepackage{academicons}
\usepackage{cite}
\usepackage[export]{adjustbox}
\usepackage{ulem}
\usepackage{svg}

\begin{document}

\title{Bangladeshi Native Vehicle Detection in Wild}

\author{Bipin Saha\textsuperscript{\orcidlink{0000-0002-3546-9074}}\and, 
        Md. Johirul Islam\textsuperscript{*\orcidlink{0000-0002-5226-0547}}\and, 
        Shaikh Khaled Mostaque\textsuperscript{*\orcidlink{0000-0003-1379-7373}} \textit{Member, IEEE}\and, 
        Aditya Bhowmik\textsuperscript{\orcidlink{0009-0006-8249-3874}}\and, 
        Tapodhir Karmakar Taton\textsuperscript{\orcidlink{0009-0000-0504-3112}}\and, 
        Md Nakib Hayat Chowdhury\textsuperscript{\orcidlink{0000-0002-1223-8556}} \and, 
        Mamun Bin Ibne Reaz\textsuperscript{*\orcidlink{0000-0002-0459-0365}} \textit{Senior Member, IEEE}  
        

\thanks{This work was supported by the Ministry of Science and Technology, Bangladesh under Reference Grant code: SRG-232388 \textit{(Corresponding
authors: Mamun Bin Ibne Reaz (mamun.reaz@iub.edu.bd); Shaikh Khaled Mostaque (misha@ru.ac.bd)
}}
\thanks{Bipin Saha is with the Department of Electrical and Electronic Engineering, University of Rajshahi, Rajshahi 6205, Bangladesh (e-mail: bipinsaha.bd@gmail.com).}
\thanks{Md. Johirul Islam is with the Department of Physics, Rajshahi University of Engineering and Technology, Rajshahi 6204, Bangladesh (e-mail: johirul@phy.ruet.ac.bd).}
\thanks{Shaikh Khaled Mostaque is with the Department of Electrical and Electronic Engineering, University of Rajshahi, Rajshahi 6205, Bangladesh (e-mail: misha@ru.ac.bd).}
\thanks{Aditya Bhowmik is with the Department of Electrical and Electronic Engineering, University of Rajshahi, Rajshahi 6205, Bangladesh (e-mail: bhowmik.aditya0@gmail.com).}
\thanks{Tapodhir Karmakar Taton is with the Department of Electrical and Electronic Engineering, University of Rajshahi, Rajshahi 6205, Bangladesh (e-mail: tapodhirtaton@gmail.com).}
\thanks{Md Nakib Hayat Chowdhury is with the Department of Electrical Electronic and Systems Engineering, Universiti Kebangsaan Malaysia, Malaysia (e-mail: nakib2025@gmail.com).}
\thanks{Mamun Bin Ibne Reaz is with the Electrical and Electronic Engineering, Independent University, Bangladesh (e-mail: mamun.reaz@iub.edu.bd).}
\thanks{}
\thanks{}
\thanks{}
\thanks{}

\thanks{This paper is submitted to IEEE Transactions on Intelligent Transportation Systems. The manuscript was written in December 2023 and revised in May 2024, and submitted in May 2024.}}


\maketitle

\begin{abstract}
The success of autonomous navigation relies on robust and precise vehicle recognition, hindered by the scarcity of region-specific vehicle detection datasets, impeding the development of context-aware systems. To advance terrestrial object detection research, this paper proposes a native vehicle detection dataset for the most commonly appeared vehicle classes in Bangladesh. 17 distinct vehicle classes have been taken into account, with fully annotated 81542 instances of 17326 images. Each image width is set to at least 1280px. The dataset’s average vehicle bounding box-to-image ratio is 4.7036. This Bangladesh Native Vehicle Dataset (BNVD) has accounted for several geographical, illumination, adverse weather conditions, variety of vehicle sizes, and orientations to be more robust on surprised scenarios. In the context of examining the BNVD dataset, this work provides a thorough assessment with four successive You Only Look Once (YOLO) models, namely YOLO v5, v6, v7, and v8. These dataset's effectiveness is methodically evaluated and contrasted with other vehicle datasets already in use. The BNVD dataset exhibits mean average precision(mAP) at 50\% intersection over union(IoU) is 0.848 corresponding precision and recall values of 0.841 and 0.774. The research findings indicate a mAP of 0.643 at an IoU range of 0.5 to 0.95. The experiments show that the BNVD dataset serves as a reliable representation of vehicle distribution and presents considerable complexities. \textcolor{cyan}{Dataset: \href{https://bipin-saha.github.io/BNVD/}{https://bipin-saha.github.io/BNVD/}}

\end{abstract}

\begin{IEEEkeywords}
Vehicle Detection, Autonomous Driving, Vehicle Dataset, Robotics, YOLO Models, Dataset Analysis.
\end{IEEEkeywords}
\section{Introduction}
\IEEEPARstart{T}{he} term “Vehicle Detection” in object detection refers to the localization and classification of targets of interest i.e. nearby or moving vehicles\cite{b1}.  The deployment of vehicle detection has been showing its potentiality in autonomous driving \cite{inital_autonomousDriving}, visual localization, pose estimation \cite{inital_Localization}, traffic management \cite{inital_TrafficManagement}, surveillance \cite{initial_survilance}, smart cities, aerial infrastructure mapping, and several unexplored areas. Accurate detection in localization is crucial for achieving the highest levels of autonomy, specifically in L4 and L5, within autonomous driving systems \cite{autonomy}. This requirement holds true even in challenging, cluttered visual environments and extreme scenarios, as highlighted in references \cite{adverse_weather}, \cite{adverse_weather4}. Needless to say, in the terms of detection, tracking, counting, and congestion management, traffic management and surveillance also play a significant role in achieving accuracy \cite{ieee_iot, ieee_iot4, ieee_iot5, ieee_iot6, adverse_weather9, adverse_weather1, adverse_weather2, adverse_weather8}. \\

Advancement of resource-hungry Convolution Neural Network (CNN) \cite{b11} and several Computer Vision techniques \cite{ieee_iot} accelerates the Transfer Learning process to work with both aerial and ground level datasets \cite{dota, dota3, dota19, dota32, dota39, bdd100k, elsevierTFlearn}. For instance, deep learning-powered traffic object recognition employing camera data yields excellent accuracy and has developed significant improvement in autonomy and self-driving applications \cite{adverse_weather8, adverse_weather9}, Although several studies have already been carried out, very few of them have been conducted in subcontinental settings, particularly in Bangladesh \cite{carld, b1, poribohonbd, dhakaai, bd1, bd2}. Most of the reprehension is caused by inadequate infrastructure of traffic signals, unmarked road signs, and uneven pedestrian crossing across the road. Edgewise, detecting native vehicles like three-wheelers, wheelbarrows, and power tillers is a bit difficult as they all  contain different  visual features, and have not  yet been trained in  any of the previous works \cite{dhakaai, b1, poribohonbd}. \\

To carry out vehicle detection research in such a hostile environment, we propose a Bangladesh Native Vehicle Detection (BNVD) dataset with 17326 vehicle images from different camera sensors and platforms. The resolution of each image is roughly 1280×720 pixels and contains vehicles of different scales, orientations, and shapes. The fully annotated BNVD dataset consists of 81542 instances of 17 different vehicle classes. In this context, the believed contributions to this research are as follows,
\begin{itemize}
    \item  So far, the proposed BNVD is the largest establishment of datasets of \textcolor{black}{View-Point Invariant Vehicle Detection} in the Bangladeshi environment, with a wide variety of classes, highly focused on native classes.
    \item  The research delved into identifying the most suitable model for vehicle detection, exploring performance fluctuations resulting from the sequential transition between different models.
    \item This study investigated how category-specific performance impacts the overall effectiveness of the detectors employed in this analysis.
    \item \textcolor{black}{Furthermore, an assessment of performance was conducted on subsets characterized by adverse weather conditions and varying light conditions. This evaluation utilized the two most effective models identified in the study.}

\end{itemize}

\section{Literature Review}

In recent years, the demand for quality datasets has increasingly grown in any kind of data-driven research\cite{itransport, dota6, dota14, dota33, dota36, dota38, dota40}. DOTA \cite{dota}, TAS \cite{dota9} , VEDAI \cite{dota25}, COWC \cite{dota24}, DLR 243K Munich Vehicle \cite{dota16} have only focused on aerial imagery. Wherein our strong focus is on street view datasets. KITTI Vision Benchmark Suite and Cityscapes 3D are considered as the most prominent datasets in autonomous driving and robotic vision. Whether KITTI uses plenty of vision, lidar sensors along with GPS and IMU mounted on a vehicle for a specific city. Cityscapes 3D dataset captures diverse data from 50 cities over several months, showcasing various seasons\cite{kitti, cityscapes}. Both of the datasets have 7 different distinct vehicle classes excluding pedestrian. Cityscapes 3D contains ground, human, and road infrastructures in their dataset, whereas KITTI gets an advantage for LiDAR data and pose odometry. In case of adverse weather conditions, DAWN dataset \cite{adverse_weather} has proposed an extraordinary workflow of visibility enhancement scheme. The dataset consists of 1K images of 7 well-known classes in real traffic scenarios. The Automatic White Balance fused by Laplacian Pyramid (AWBLP) lessens visibility limits in extreme weather situations such as fog, rain, snow, and sand. It has been tested with 21 state-of-the-art object detection models. \textcolor{black}{Richly annotated driving videos make up the BDD100K dataset, which is widely used for training and evaluating deep learning algorithms in autonomous driving applications such as object recognition, scene segmentation, and lane detection. It is made up of 100,000 films that were shot in various US sites under various weather and environmental situations \cite{bdd100k}. \textcolor{black}{The CityFlow dataset serves as a comprehensive city-scale benchmark tailored for tracking and re-identifying vehicles across multiple targets and cameras. It includes more than 3 hours of synchronized high-definition videos captured by 40 cameras situated at 10 different intersections within a mid-sized U.S. city. Additionally, the dataset offers camera geometry and calibration details to support spatio-temporal analysis \cite{cityflow}}. Prioritizing the concern of occlusion handling, the Rope3D dataset includes both frontal view and roadside camera images to address the limitations of existing perception datasets for autonomous driving \cite{rope3d}. }

The above studies hardly match with South Asian scenario. \textcolor{black}{The transportation system in Bangladesh grapples with significant deficiencies, particularly in its road infrastructure, which is often inadequate in size, poorly maintained, and lacking in resources for upkeep \cite{roadcondition1}. Factors such as high traffic density, substantial rainfall leading to waterlogging, inconsistent road markings, signage, and lane divisions pose considerable difficulties \cite{roadcondition2}. Moreover, the presence of informal settlements nestled amidst roads adds further complexity to the overall landscape of these transportation networks.} \textcolor{black}{There are two closest match IDD and CARL-D datasets. IDD contains 10,004 images from 182 drive sequences in Bangalore and Hyderabad, India, with an average resolution of 1678x968 pixels. It features various adverse weather conditions, including heavy shadows, cloudy skies, and variations in particulate matter from fog, dust, or smog. The images capture different lighting conditions, times of day, and weather scenarios \cite{idd}.} While CARL-D \cite{carld} authors drive around 1000 KM highway, through 100+ cities in 3 provinces of Pakistan. Here, the dataset consists of 25 distinct classes including vehicles and road signs. Though Indian and Pakistani roads are almost similar to the Bangladeshi environment, there are dissimilarities in terms of local vehicles. In Bangladesh, only a few vehicle recognition works \cite{b1, poribohonbd, dhakaai, p2dhaka} have been performed with a low variance of images. PoribohonBD \cite{poribohonbd} and Dhaka-AI \cite{dhakaai} are the two most buzzworthy datasets that are available considering the Bangladeshi scenario. Of them, PoribohonBD \cite{poribohonbd} consists of 16 Classes, with 26851 vehicle instances of 9058 images. Though the particular dataset primarily focuses on road vehicles, it also considers water transport with Boat and Launches. The authors proposed a data augmentation technique to address the challenges of a limited number of images and class imbalance, aiming to enhance the model's accuracy.  Besides Dhaka-AI \cite{dhakaai} has a larger number of classes with a comparatively lower number of images and instances. It is confronted with a substantial number of class imbalance issues. In a few classes e.g. minibus, ambulance, taxi, army vehicle, scooter, police car, and garbage van, the dataset has only 95, 70, 60, 43, 38, 32, and 3 instances which may lead to limited object representation and biased or imbalanced training. In a broad sense, the dataset may suffer from limited robustness and variations. Another dataset named P2 Dhaka \cite{p2dhaka} has a better representation than Dhaka-AI but the number of classes is comparatively less than the available datasets. A brief comparison among datasets has been given in Table 1 and Fig-1.



\begin{table*}[htbp]
\centering
\caption{Vehicle Detection Dataset Comparison (Horizontal Bounding Box)}
\begin{tabular}{|c|c|c|c|c|c|cccc|}
\hline
{Dataset} & {Categories} & {\begin{tabular}[c]{@{}c@{}}Instances \\ (BBox)\end{tabular}} & {Images} & {Image Width} & {\begin{tabular}[c]{@{}c@{}}Average \\ BBox/Image\end{tabular}} & \multicolumn{4}{c|}{Diversity}                                                                                   \\ \cline{7-10} 
                         &                             &                                                                              &                         &                              &                                                                                & \multicolumn{1}{c|}{Rainy}      & \multicolumn{1}{c|}{Foggy}      & \multicolumn{1}{c|}{Night}      & Limited    \\ \hline
KITTI      \cite{kitti}              & 8                           & 80000                                                                        & 15000                   & 1392px                       & 5.3333                                                                         & \multicolumn{1}{c|}{-}          & \multicolumn{1}{c|}{-}          & \multicolumn{1}{c|}{-}          & -          \\ \hline
CityScapes 3D    \cite{cityscapes}        & 8                           & 27000                                                                        & 5000                    & 2048px                       & 5.4                                                                            & \multicolumn{1}{c|}{-}          & \multicolumn{1}{c|}{-}          & \multicolumn{1}{c|}{-}          & -          \\ \hline
Rope3D     \cite{rope3d}              & 12                          & 1500000                                                                      & 50000                   & 1920px                       & 30                                                                             & \multicolumn{1}{c|}{Y}          & \multicolumn{1}{c|}{-}          & \multicolumn{1}{c|}{Y}          & Y          \\ \hline
SEU\_PML     \cite{seu_pml}            & 13                          & 270684                                                                       & 6588                    & 1920px to 4096px             & 41.0874                                                                        & \multicolumn{1}{c|}{Y}          & \multicolumn{1}{c|}{-}          & \multicolumn{1}{c|}{Y}          & Y          \\ \hline
BDD100K      \cite{bdd100k}            & 12                          & 2221128                                                                      & 100000                  & 1280px                       & 22.2112                                                                        & \multicolumn{1}{c|}{Y}          & \multicolumn{1}{c|}{Y}          & \multicolumn{1}{c|}{Y}          & Y          \\ \hline
CityFlow     \cite{cityflow}            & -                           & 229680                                                                       & $\sim$666               & 960px                        & 344.9                                                                          & \multicolumn{1}{c|}{-}          & \multicolumn{1}{c|}{Y}          & \multicolumn{1}{c|}{-}          & Y          \\ \hline
IDD        \cite{idd}              & 34                          & 77142                                                                        & 10004                   & 1678px                       & 7.711                                                                          & \multicolumn{1}{c|}{-}          & \multicolumn{1}{c|}{Y}          & \multicolumn{1}{c|}{Y}          & Y          \\ \hline
CARL-D     \cite{carld}              & 25                          & 50348                                                                        & 15000                   & -                            & 3.3565                                                                         & \multicolumn{1}{c|}{-}          & \multicolumn{1}{c|}{-}          & \multicolumn{1}{c|}{-}          & Y          \\ \hline
DhakaAI       \cite{dhakaai}           & 21                          & 24368                                                                        & 3003                    & -                            & 8.1145                                                                         & \multicolumn{1}{c|}{-}          & \multicolumn{1}{c|}{-}          & \multicolumn{1}{c|}{Y}          & Y          \\ \hline
P2 Dhaka      \cite{p2dhaka}           & 8                           & 43796                                                                        & 4777                    & -                            & 9.1680                                                                         & \multicolumn{1}{c|}{-}          & \multicolumn{1}{c|}{Y}          & \multicolumn{1}{c|}{-}          & -          \\ \hline
PoribohonBD     \cite{poribohonbd}        & 16                          & 26851                                                                        & 9058                    & -                            & 2.9643                                                                         & \multicolumn{1}{c|}{Y}          & \multicolumn{1}{c|}{-}          & \multicolumn{1}{c|}{Y}          & Y          \\ \hline
\textbf{BNVD}            & \textbf{17}                 & \textbf{81542}                                                               & \textbf{17326}          & \textbf{1280px}              & \textbf{4.7036}                                                                & \multicolumn{1}{c|}{\textbf{Y}} & \multicolumn{1}{c|}{\textbf{Y}} & \multicolumn{1}{c|}{\textbf{Y}} & \textbf{Y} \\ \hline
\end{tabular}
\end{table*}

Eliminating bias over training and better data representation for low-instance classes, data augmentation is a popular technique. The advancement of the Generative Adversarial Network (GAN) creates another branch of synthetic image data generation.  \cite{yolov4traffic}. The majority of datasets have hardly seen night images, for this style transfer technique via GAN has been suggested for recent years \cite{styletransfer}. One significant limitation has been observed in low light conditions, where the camera shutter speed decreases, resulting in motion blur for fast-moving vehicles. This particular aspect is comparatively very hard to effectively replicate using style transfer approaches. Compared to the aforementioned datasets, the proposed BNVD is more applicable to real-world scenarios as it includes tremendous vehicle instances. 

\section{Annotation of BNVD}

A robust dataset enables models to handle diverse scenarios by encompassing a wide range of objects, backgrounds, and illumination conditions. Precise identification is ensured through a variety of categories, sizes, and positions. Consistent, annotated data remains essential for efficient model training. Each class contains an adequate number of images for optimal training. Avoiding bias is achieved by equally spreading data across categories, while real-world images ensure practical usage. According to these criteria, annotation of BNVD is as follows,

\subsection{Image Collection}
  Image resolution and variation in sensor size a vital factors in the object detection pipeline \cite{dota5}.  To eliminate sensor biases and enhance robustness we employ a variety of cameras including mobile phones (Google Pixel 6, 50MP; Xiaomi Pocophone F1, 12MP; Redmi K20 Pro, 48MP; Realme 6, 64MP), action cameras (GoPro Hero 9 Black, 20MP) and YouTube videos having multiple resolution and orientation considering a variety of lighting scenarios.

\begin{figure}[!t]
\centering
\includegraphics[width=3.5in]{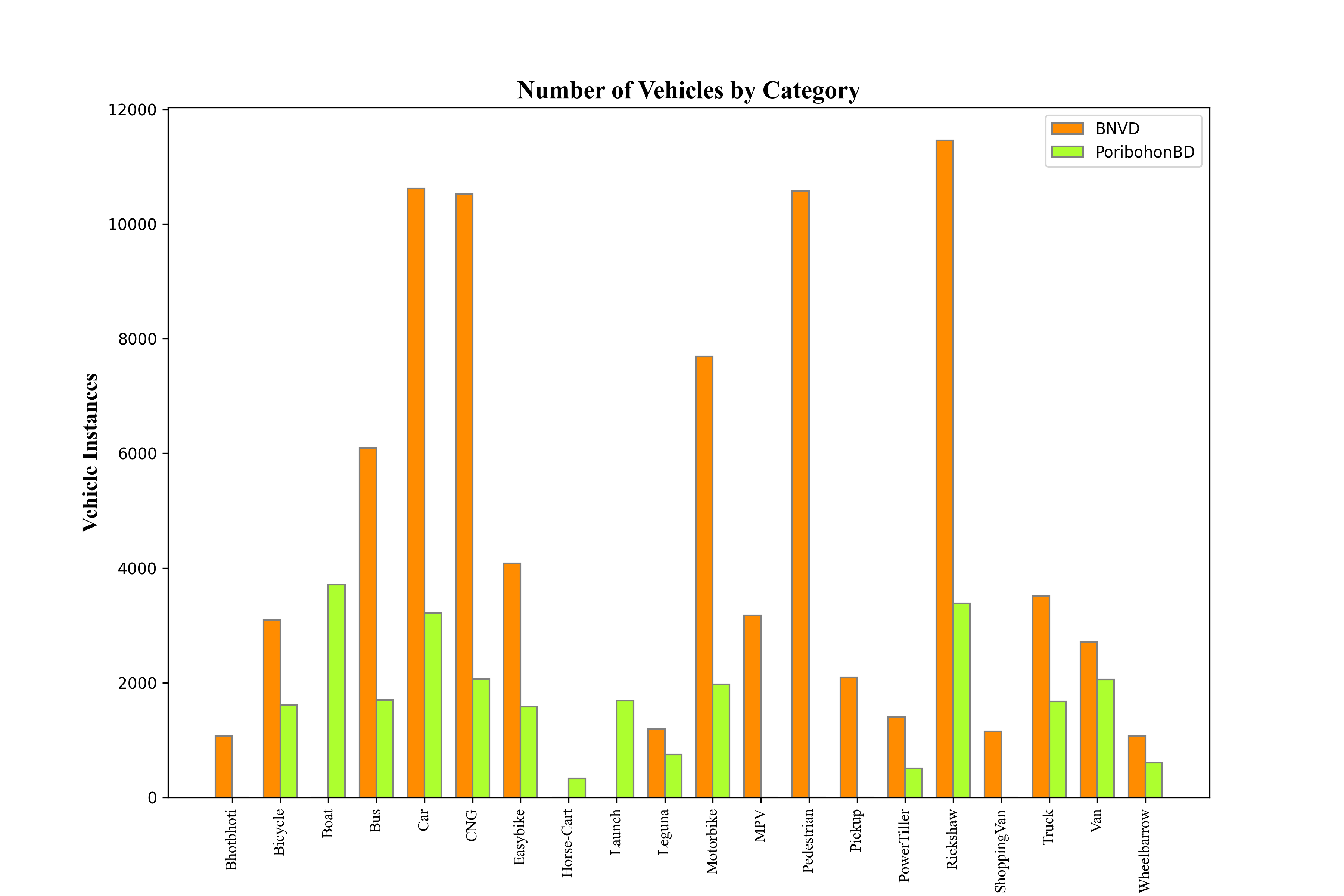}
\caption{Vehicle category instances comparison between PoribohonBD and BNVD dataset}
\label{fig_1}
\end{figure}

\subsection{Category Selection}
\begin{figure*}[!t]
  \centering
  \includegraphics[scale=1.5]{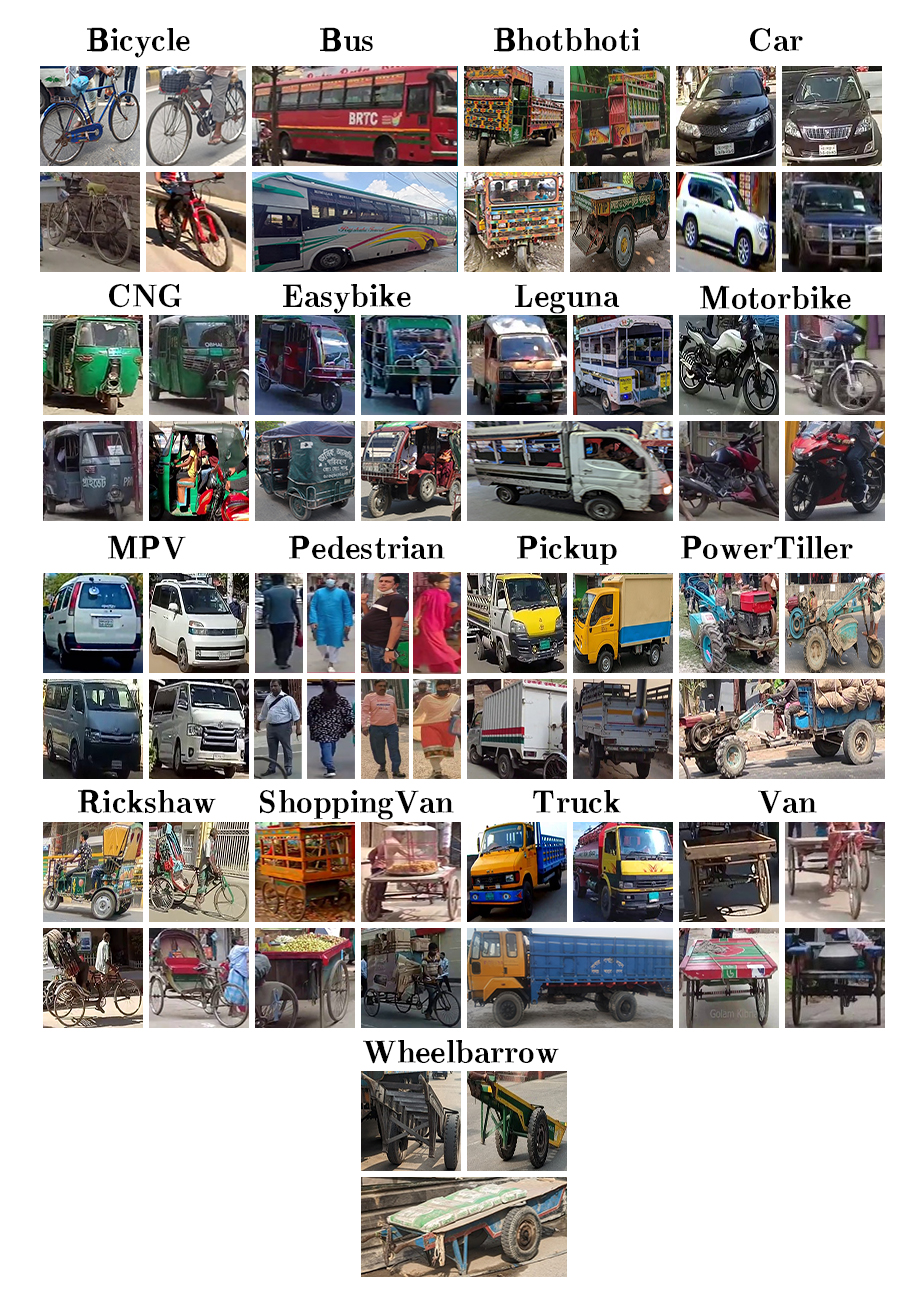} 
  \caption{Overview of Bangladeshi Native Vehicle Dataset (BNVD)}
  \label{fig:2}
\end{figure*}
The BNVD dataset consists of a variety of vehicle groups, spanning Bicycle, Bus, Bhotbhoti(single cylinder diesel engine powered local three-wheeler), Car, CNG (Compressed natural gas-powered three-wheeler), Easybike, Leguna, Motorbike, MPV (Multi-Purpose Vehicle), Pedestrian, Pickup, Power Tiller, Rickshaw, Shopping Van, Truck, Van, and Wheelbarrow.\\
The categories have been selected based on visual appearance and unique features among them. Though similar kinds of vehicles are put in their allocated category, for some classes, they are differentiated by their visual structure and appearance. For instance, Fig-3 represents rickshaws with 3 different structures that are visually separated from one another.\\
\begin{figure}[!t]
\centering
\includegraphics[width=3.5in]{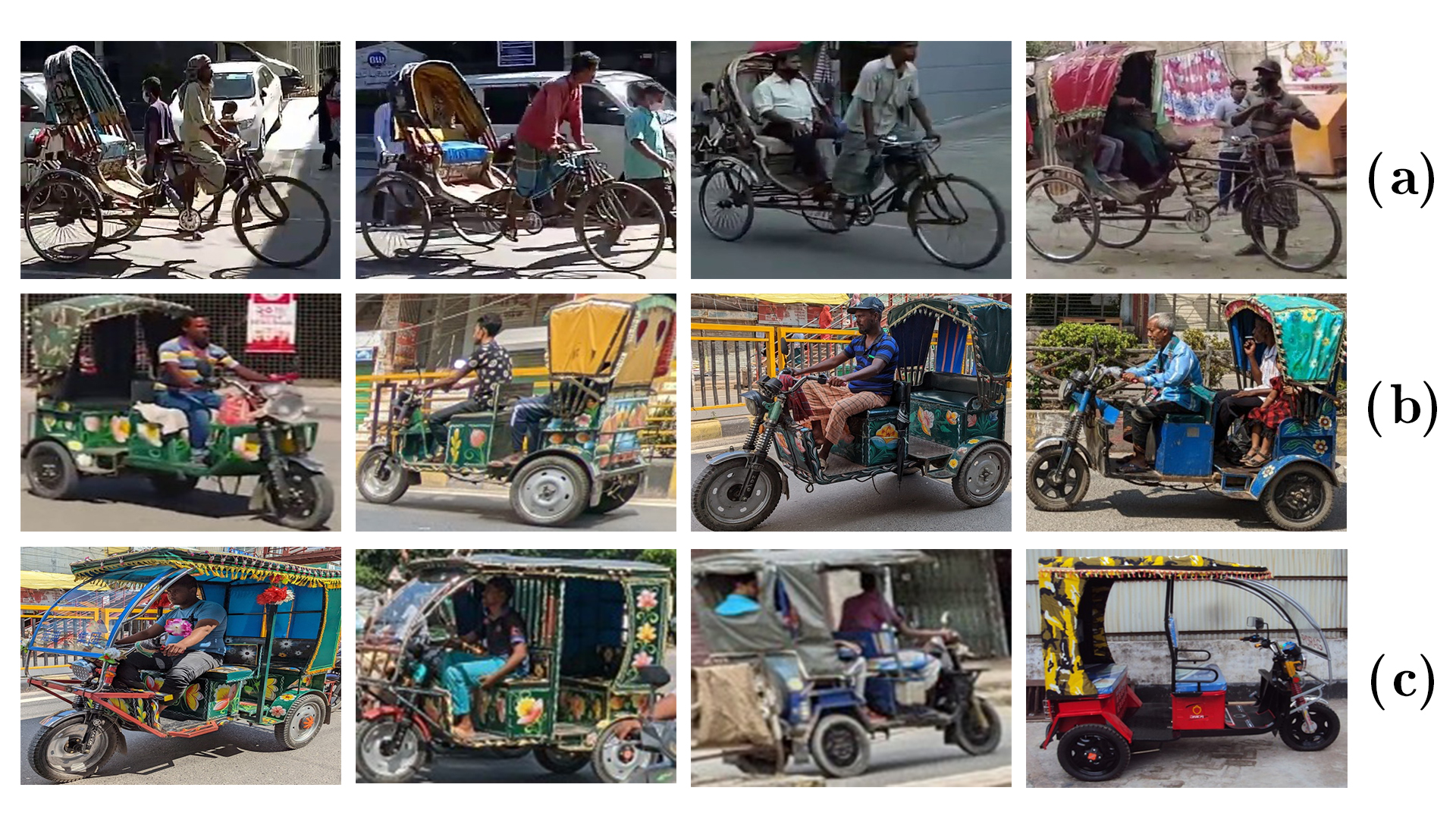}
\caption{Types of Rickshaw}
\label{fig_3}
\end{figure}
The majority of vehicles on Bangladeshi roadways are three-wheelers, such as Rickshaw, Van, Easybike, and Bhotbhoties. Regardless of sharing almost similar outward appearance, Shopping Vans and Vans serve distinct purposes. Shopping vans are generally used for selling goods while vans are designed for transporting passengers and goods. Bhotbhoties have different visual structures than other three-wheelers. The majority of them are heavily utilized in the northern region of Bangladesh for the transportation of agricultural goods or carrying passengers over a short distance. Walking passengers are common on streets, roadways, and even on highways. Although \cite{kitti, cityscapes} addresses Pedestrians, in their dataset, we hardly found any literature on sub-continental locations that deal with pedestrians. Here, the proposed incorporation of Pedestrians as a means of detecting roadside anomalies in this scenario.
\begin{figure}[!t]
\centering
\includegraphics[width=3.0in]{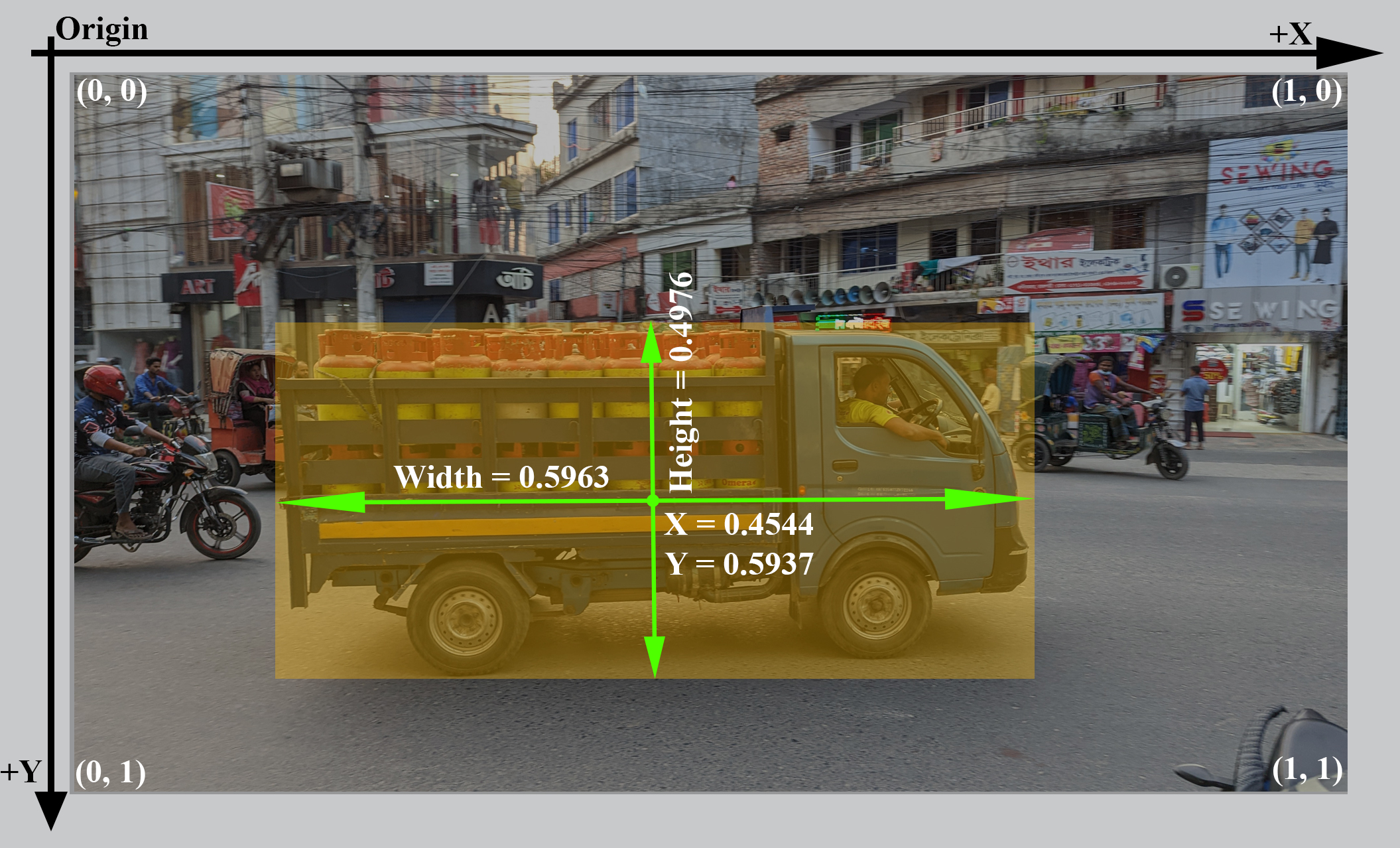}
\caption{Annotation Method}
\label{fig_4}
\end{figure}
\subsection{Annotation Method}
We utilize text annotation format instead of COCO or Pascal VOC format. In the field of computer vision, many visual concepts, such as region narratives, objects, characteristics, and connections, can be annotated with bounding boxes, as demonstrated in \cite{dota12}. In the text annotation format the bounding box coordinates are, (x, y, w, h), where (x, y) denotes the center and (w, h) denotes the bounding box's width as well as its height, correspondingly. Fig 4. represents a visual overview.

\definecolor{capri}{rgb}{0.0, 0.75, 1.0}

\subsection{Data Split}
To maintain a balance between the training and test data distributions, a random selection method has been adopted, where 70 percent of the original images are chosen in the training set, while the remaining 30 percent is divided between the validation and testing sets. The public dataset and fine tuned weights are accessible from this \href{https://www.kaggle.com/datasets/sahabipin/bangladeshi-native-vehicle-dataset}{\textcolor{capri}{website}} and \href{https://github.com/bipin-saha/BNVD}{\textcolor{capri}{GitHub}}.

\section{Properties of BNVD}
\subsection{Image Collection Area}
Though Bangladesh has less geographic diversity, there has been an indubitable difference in vehicle choice between rural and urban areas. Local transportation options have traditionally been more prevalent in rural areas, while urban settings tend to favor motorized vehicles. Prioritization has been done in both areas, and images have been gathered from 17 different districts of Bangladesh. Fig 5 illustrates the areas where the images were collected.

\begin{figure}[!t]
\centering
\includegraphics[height=2in]{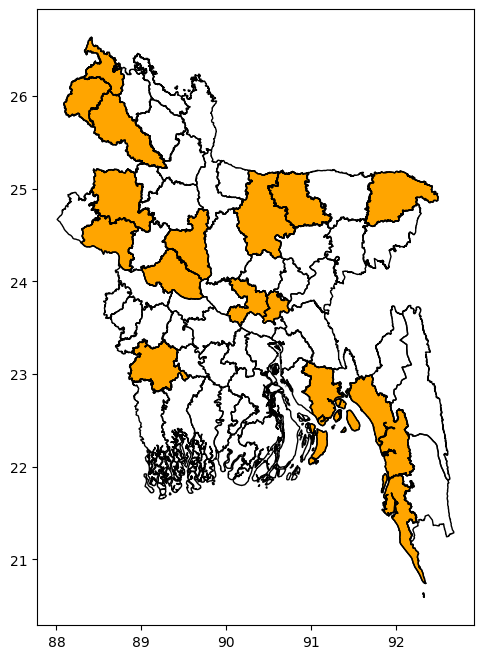}
\caption{Image Collection area of Bangladesh}
\label{fig_5}
\end{figure}

\subsection{Image Size}
In typical cases, the size of images can vary due to factors such as the camera settings, different angles, and positioning during capture. In our dataset, the majority of images have dimensions around 1280 pixels. During the annotation process, consideration is given to the original full-size images without partitioning. The approach has been adopted to keep instances intact in images. Partitioning images may result in a single instance being separated into several pieces, complicating the further annotation process and data analysis.

\subsection{Lighting Condition}
The BNVD dataset (Fig. 6) amalgamates an array of lighting conditions, encompassing optimal illumination under sunny skies, varying to subdued lighting during cloudy days and evenings, and extending to low-light scenarios at night. This compilation offers a comprehensive representation of diverse lighting environments, crucial for capturing images with enhanced detail, sharpness, color fidelity, and reduction in noise levels.

\begin{figure*}[!t]
  \centering
  \includegraphics[width=1\textwidth]{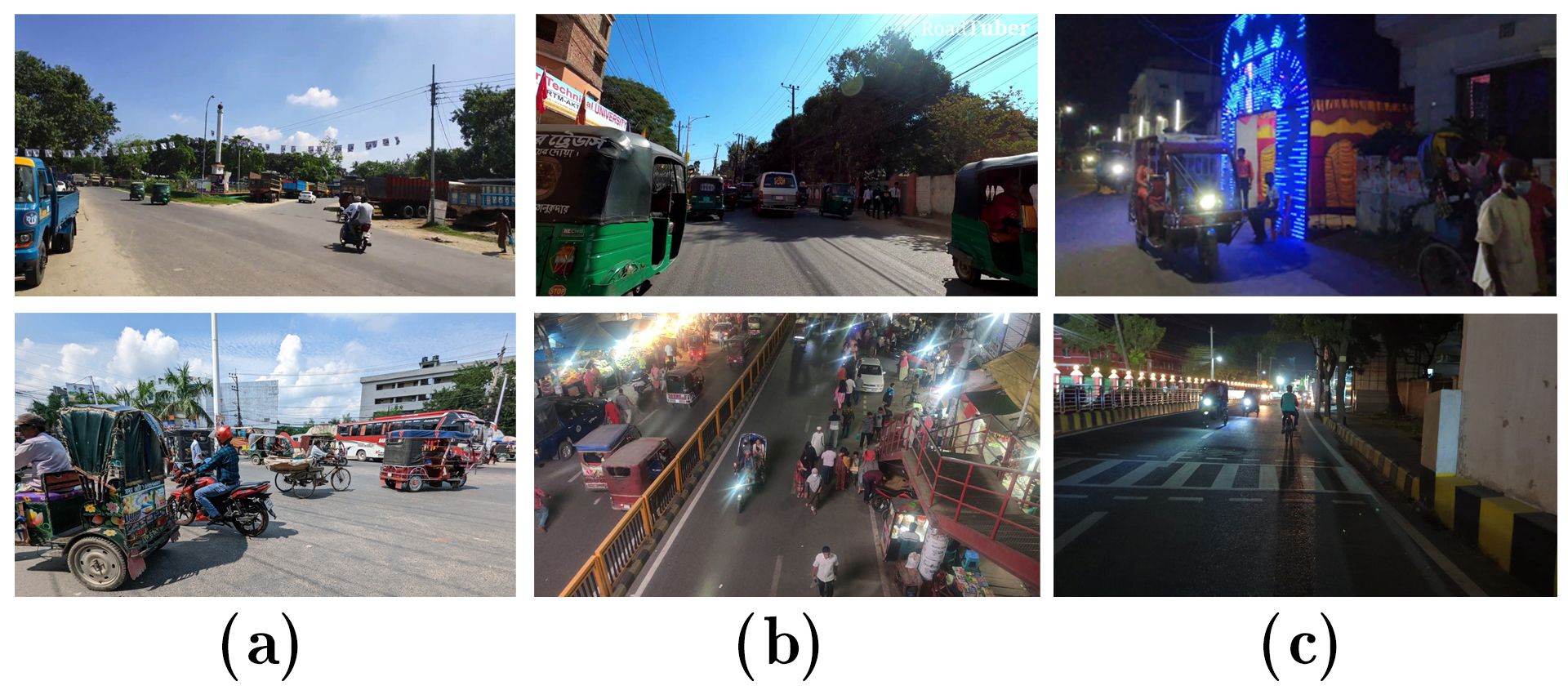} 
  \caption{Comparisons of Lighting Conditions (a) Good, (b) Limited, (c) Low}
  \label{fig:6}
\end{figure*}

\subsection{Adverse Weather Condition}
\textcolor{black}{In consideration of the prevailing weather patterns in Bangladesh, characterized by temperature conditions, the country primarily encounters adverse weather scenarios such as rain and fog. To enhance the efficacy of our dataset in capturing these conditions, we enriched it with instances of rainy and foggy weather conditions. Enriching the dataset with rainy and foggy weather instances enables robust modeling of adverse weather scenarios, improving predictive accuracy and real-world applicability. Subsequently, we conducted a comparative analysis of vehicle detection under adverse weather conditions using two distinct approaches.}

\subsection{Orientation of Instances}
The dataset exhibits a well-rounded balance in vehicle orientation, encompassing a comprehensive range of poses, including frontal, rear, side, and traffic monitoring perspectives. This is crucial for training a robust vehicle detector that can recognize objects in all kinds of orientations and perspectives. Given the diverse orientations of vehicles in real-world scenarios, detectors must effectively handle this variability to ensure comprehensive detection. 
\subsection{Spatial Resolution Information}
The inclusion of spatial resolution information in the BNVD dataset is valuable for several reasons. Primarily, it helps to address potential confusion that may arise when the same vehicle categories are captured from different distances, which may appear different in size.
 By considering spatial sampling, models are more adaptable and can focus on the geometric features of a vehicle rather than relying solely on its size, which enhances its ability to accurately classify objects, even if they are of different proportions.
Secondly, spatial resolution information proves beneficial for fine-grained classification tasks. For instance, it enables the model to distinguish between small pickups and large trucks more effectively. The details provided by spatial resolution assist in making subtle distinctions between similar categories.
Furthermore, spatial resolution information can be utilized to identify mislabeled outliers in our dataset. Since most categories exhibit limited intra-class variations in terms of actual sizes, outliers can be identified by examining objects whose size significantly deviates from others within a small range of spatial resolution. This approach aids in filtering out anomalies and enhancing the quality and consistency of the dataset.

\subsection{Pixel Size of Categories}
\begin{table}[htbp]
  \caption{\textcolor{black}{Category-wise Percentage of Object Sizes}}
  \label{tab:object_sizes}
  \begin{tabular}{lccc}
    \hline
    \textbf{Category} & \textbf{10-50px (\%)} & \textbf{50-300px (\%)} & \textbf{Above 300px (\%)} \\
    \hline
    Bicycle        & 0.1306 & 0.7909 & 0.0785 \\
    Bus            & 0.2324 & 0.6663 & 0.1013 \\
    Bhotbhoti      & 0.0511 & 0.4958 & 0.4531 \\
    Car            & 0.3224 & 0.6341 & 0.0435 \\
    CNG            & 0.2142 & 0.6991 & 0.0866 \\
    Easybike       & 0.0729 & 0.8086 & 0.1185 \\
    Leguna         & 0.0185 & 0.4404 & 0.5411 \\
    Motorbike      & 0.2144 & 0.7210 & 0.0647 \\
    MPV            & 0.1734 & 0.7214 & 0.1051 \\
    Pedestrian     & 0.0677 & 0.8320 & 0.1004 \\
    Pickup         & 0.1057 & 0.7303 & 0.1640 \\
    PowerTiller    & 0.0185 & 0.6828 & 0.2988 \\
    Rickshaw       & 0.0812 & 0.7855 & 0.1333 \\
    ShoppingVan    & 0.0400 & 0.7452 & 0.2148 \\
    Truck          & 0.1072 & 0.6780 & 0.2147 \\
    Van            & 0.2047 & 0.7478 & 0.0475 \\
    Wheelbarrow    & 0.0765 & 0.6968 & 0.2267 \\
    \textbf{BNVD Total}     & \textbf{0.1254} & \textbf{0.6986} & \textbf{0.1760} \\
    \hline
  \end{tabular}
\end{table}
According to literature suggestion \cite{dota, dota35}, the term Size of Pixels refers to the height of the bounding box, which serves as a measure of instance size. All instances are categorized into three groups based on horizontal bounding box height.
\\1. Small: Instances with heights ranging from 10 to 50 pixels
\\2. Medium: Instances with heights ranging from 50 to 300 pixels.
\\3. Large:  Heights above 300 pixels.\\ 
Table II is a class-wise representation of vehicle instance sizes in percentage. The instance size is highly distributed in the medium category, while a good balance in the small and large categories.
\subsection{Instance Density of Image}
\begin{figure}[!t]
\centering
\includegraphics[width=3in]{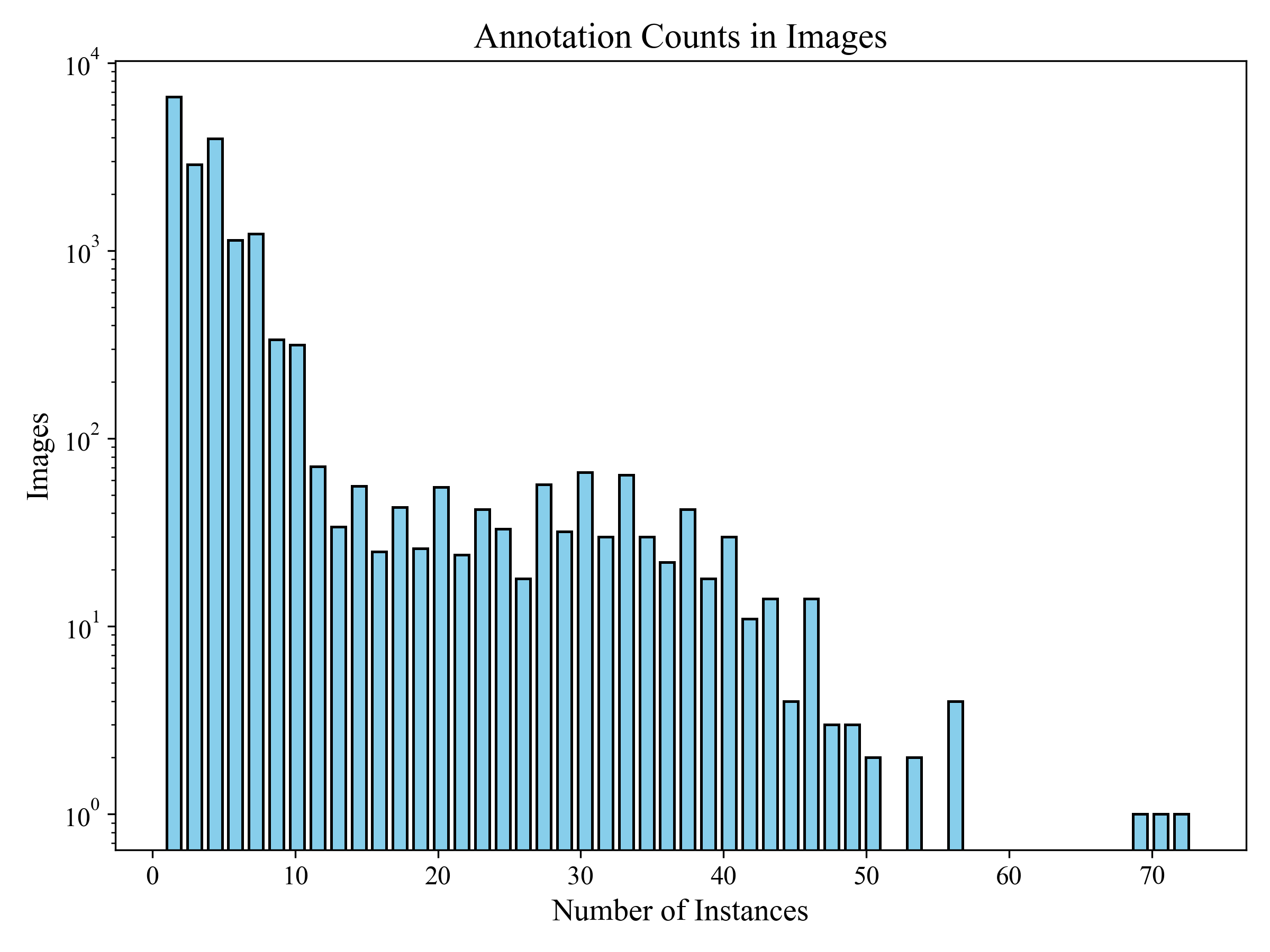}
\caption{Histogram of the number of annotated instances per image.}
\label{fig_7}
\end{figure}
Fig-7 represents the annotation per image histogram. The histogram graph demonstrates that there is at least a single annotation present in all of the images in the dataset. Moreover, a huge number of images have 1 to 5 instances. Comparatively highest number of instances for an image is 73.

\section{Model Selection}
The evaluation of the CNN and Transformer algorithms has replaced traditional object detection approaches. At CNN the traditional workflow involves extracting high-level feature maps from the input and categorizing them by locating objects within images. Over time it became a standard tool for many Computer Vision and machine learning applications. To match accuracy and real-time performance, computer vision algorithms have developed in two directions. 'Two-Stage' and 'Single Stage' object detectors, where one focused on accuracy and another one focused on speed. Two-stage detectors are highly region-based, with computational complexity.  This work focuses on single-stage object detection algorithms known for their speed and rational accuracy. One-stage detectors immediately identify the class of objects and their coordinate offsets rather than detecting object versus non-object, as it is done in the RPN network of the region-based architecture. Hence, they are also known as regression-based detectors \cite{ieee_iot}. 
\begin{table}[]
\caption{Two Stage and Single Stage Model Performance Comparison on COCO Dataset}
\begin{tabular}{|c|c|c|c|c|}
\hline
Stage                   & Model         & Backbone                                          & AP   & FPS  \\ \hline
{}{}{Two}      & Fast R-CNN \cite{ieee_iot}        & {}{}{VGG-16}         & 19.7 & 0.5  \\ \cline{2-2} \cline{4-5} 
                        & Faster R-CNN \cite{ieee_iot}      &                               & 21.9 & 5    \\ \cline{2-5} 
                        & TridentNet \cite{ieee_iot}        & ResNet-101                    & 48.4 & -    \\ \hline
{}{}{Single}   & SSD \cite{ieee_iot}               & VGG-16                        & 33.2 & 6.4  \\ \cline{2-5} 
                        & RetinaNet \cite{ieee_iot}         & ResNet-101 FPN                & 37.8 & 5.05 \\ \cline{2-5} 
                        & Efficient Det \cite{ieee_iot}     & EfficientNet                  & 55.1 & 3.5  \\  \hline 
                        & YOLOv5 \cite{git_yolov5}          & Modified CSPNet               & 45.4 & 182  \\ \cline{2-2} \cline{4-5} 
                        & YOLOv6 \cite{git_yolov6}          &                               & 50.0 & 175  \\ \cline{2-2} \cline{4-5} 
                        & YOLOv7 \cite{git_yolov7}          &                               & 51.4 & 161  \\ \cline{2-2} \cline{4-5} 
                        & YOLOv8 \cite{git_yolov8}          &                               & 50.2 & 546  \\ \hline
\end{tabular}
\end{table}Among single-stage detectors YOLO is a real-time object detection algorithm developed by Joseph Redmon in 2015 \cite{yolov1}. It addresses object detection as a single-pass regression problem and is able to detect multiple objects in a single forward pass through the neural network. By simultaneously identifying every bounding box, YOLO unified the object detection procedures. In order to do this, YOLO splits the input image into a S × S grid and forecasts B bounding boxes belonging to the same class, as well as its confidence for C distinct classes for each grid element. Pc, bx, by, bh, and bw are the five values that make up each bounding box prediction. Pc is the box's confidence score, which indicates the accuracy and degree of the model's belief that the box includes an item. The complete image's height and width are represented by the coordinates bh and bw, whereas the box's centers in relation to the grid cell are denoted by bx and by. After using non-maximum suppression (NMS) to eliminate duplicate detections, the output of YOLO is a tensor of S × S × (B × 5 + C). Table III shows how the YOLO family is dominating the single-stage object detection community due to its high degree of accuracy, inference speed, fit with industrial demands, including edge-friendly deployment conditions, precision, and lightweight design. Hence, the focus remained solely on four successive cutting-edge models within the You Only Look Once (YOLO) framework. They are YOLO v5 \cite{git_yolov5}, v6\cite{git_yolov6}, v7\cite{git_yolov7} and v8\cite{git_yolov8}. 

Table IV provides a detailed explanation of the motivation for selecting these models. Though each model displays better performance in the COCO dataset while increasing model complexity with the number of parameters, there is higher interference time in terms of real-time applications. To achieve a better trade-off between inference speed and performance, such factors motivate us to stay with the medium-sized model solely.

\begin{table*}[htbp]
\caption{YOLO Model Comparison}
\centering
\begin{tabular}{|c|c|p{7cm}|c|c|c|}
\hline
Model & Parameters(M) & Features & Backbone & mAP0.5:0.95& Year\\
\hline

 {}{}{} &  & 1. YOLOv8 utilizes a similar backbone to YOLOv5 but with changes to the CSPLayer (C2f module). The C2f module combines high-level features with contextual information, enhancing detection accuracy \cite{arxivyolo}.  &  & & \\
 {}{}{YOLOv8} & 25.9 & 2.YOLOv8 employs an anchor-free model with a decoupled head to independently handle objectness, classification, and regression tasks. This design enhances overall accuracy. In the output layer, the sigmoid function is used for the objectness score, representing the probability of containing an object, while the softmax function is used for class probabilities \cite{arxivyolo}.  & Modified CSPNet & {50.2} & Jan, 2023 \\
 {}{}{\cite{git_yolov8}} & {(v8m)} & 3.YOLOv8 implements CIoU and DFL loss functions for bounding box loss, along with binary cross-entropy for classification loss. These loss functions improve object detection performance, especially for smaller objects \cite{arxivyolo}.  & &  & \\
\hline

 {}{}{} &  & 1. YOLO v7 introduces a series of architectural reforms aimed at enhancing accuracy while maintaining a high detection speed \cite{mdpiYolov1t8}.  &  & & \\
 {}{}{YOLOv7} & {36.9} & 2.The Extended Efficient Layer Aggregation Networks (E-ELAN) are introduced in YOLO v7, utilizing a strategy that prioritizes learning at different depths for improved convergence and learning efficiency. This approach enhances model accuracy without increasing complexity or computational resources \cite{mdpiYolov1t8, mdpiSensorsYOLO}.  & Modified CSPNet & {51.4} & July, 2022 \\
 {}{}{\cite{git_yolov7}} & {(v7)} & 3.Scaling for Concatenation-based Models (MSCM) in YOLO v7 adjusts the number of stages. By scaling the depth of a computational block while maintaining corresponding width scaling in the transmission layer, the input width of subsequent layers is increased, enhancing model performance \cite{mdpiYolov1t8, mdpiSensorsYOLO}.  & &  & \\
 {}{}{} &  & 4.In YOLOv7, re-parameterization is performed to combine it with a different network, resulting in a 40\% reduction in parameters and a 50\% reduction in computation for the object detector \cite{mdpiSensorsYOLO}.  & &  & \\
\hline 

 {}{}{} &  & 1.YOLO-v6 takes an anchor-free approach, making it 51\% faster than anchor-based methods \cite{mdpiYolov1t8}.  & &  & \\
 {}{}{YOLOv6} & {34.9} & 2.YOLO-v6 utilizes a decoupled head architecture with extra layers that enhance performance by separating features from the final head \cite{mdpiYolov1t8}.  & Modified CSPNet & {50.0} & June 2022 \\
 {}{}{\cite{git_yolov6}} & {(v6m)} & 3.Incorporating Varifocal loss (VFL) for classification and distribution focal loss (DFL) for regression, resulting in performance improvement \cite{mdpiYolov1t8}. & &  & \\
\hline
 {}{}{} &  & 1.The backbone features a modified CSPDarknet53 with a streamlined Stem, utilizing a large-window strided convolution layer to optimize memory and computation. Subsequent convolutions extract key image features \cite{arxivyolo}. & &  & \\
 {}{}{YOLOv5} & {21.2} & 2.The SPPF layer and subsequent convolutions handle features at multiple scales, while the upsample layers enhance feature map resolution. SPPF efficiently combines different scale features into a fixed-size map, accelerating network computation \cite{arxivyolo}. & Modified CSPNet & {45.4} & June, 2020 \\
 {}{}{\cite{git_yolov5}}  & {(v5m)}  & 3.YOLOv5 incorporates diverse augmentations like Mosaic, copy-paste, random affine, MixUp, HSV augmentation, random horizontal flip, and albumentations package augmentations. It enhances grid sensitivity for better stability against runaway gradients \cite{arxivyolo}. & &  & \\
 {}{}{} &  & 4.PyTorch framework is easy to train than the Darknet framework which is used in YOLO V4 \cite{mdpiYolov1t8}.& &  & \\
\hline
\end{tabular}
\end{table*}

\section{Experimental Analysis}
\subsection{Evaluation Metric}
Evaluation metrics are quantitative measures used to assess the performance or quality of a system, model, or algorithm. They provide objective criteria for comparing different approaches and are crucial in understanding the system's strengths and weaknesses. Common evaluation metrics include Average Precision (AP), Mean Average Precision (mAP), Intersection over Union (IoU), Precision, and Recall.\\
\textbf{Average Precision (AP)}: Average Precision is the metric of evaluating the performance of an information retrieval or object detection system. AP measures the precision-recall trade-off and calculates the area under the precision-recall curve. AP summarizes the algorithm's performance in detecting objects across various confidence thresholds. A higher value of AP indicates better performance \cite{b1}. \\

\textbf{Mean Average Precision (mAP)}: Mean Average Precision is the average of AP values across multiple object categories. It provides an overall measure of performance for object detection algorithms. Each class is evaluated independently, and mAP is computed by averaging the AP values across all classes. Better performance is indicated by higher values of mAP\cite{b1}. \\
\begin{equation*}
mAP = \frac{1}{N} \sum_{i=1}^{k=N} AP_k
\end{equation*}
Here, N = number of classes, AP = Average Precision over class k\\

\textbf{Intersection over Union (IoU)}: IoU measures the overlap between predicted bounding boxes and ground truth bounding boxes. It is used as a threshold for determining whether a predicted bounding box is considered a true positive. At the calculation, the ratio of the area of intersection between the predicted and ground truth regions to the area of their union. IoU ranges from 0 to 1, where a value of 1 indicates a perfect overlap between the predicted and ground truth regions. IoU helps assess the accuracy of object localization \cite{b1}. \\
\begin{equation*}
    IoU = \frac{\textit{Area of Overlap}}{\textit{Area of Union}}
\end{equation*}
\textbf{Precision and Recall}: Precision measures the proportion of correctly detected objects among the predicted objects, while recall measures the proportion of correctly detected objects among the actual objects. These metrics provide insights into the algorithm's ability to accurately identify objects.\\
\begin{equation*}
    Precision = \frac{\textit{True Positive}}{\textit{True Positive + False Positive}}
\end{equation*}
\begin{equation*}
    Recall = \frac{\textit{True Positive}}{\textit{True Positive + False Negative}}
\end{equation*}
In evaluating vehicle detection performance on datasets, the focus centers on performance assessing with state-of-the-art algorithms. Specifically, the benchmark algorithms selected include YOLOv5, YOLOv6, YOLOv7, and YOLOv8. These algorithms have been chosen based on their exceptional performance in general object detection tasks as mentioned in Table III and IV. Model hyperparameters have been kept as their default setting as mentioned in Table V. For the model, YOLO v6, “yolov6-m finetuned” has slightly different baseline hyperparameter settings in terms of IoU, Momentum, Weight Decay, Learning rate scheduling, Warmup Momentum, and Warmup Bias Learning Rate.\\

\begin{table}[htbp]
  \caption{Category-wise Percentage of Object Sizes}
  \label{tab:object_sizes_percentage}
  \begin{tabular}{ccccc}
    \hline
    \textbf{Parameter} & \textbf{YOLOV5} & \textbf{YOLOV6} & \textbf{YOLOV7} & \textbf{YOLOV8} \\
    \hline
    Optimizer        & SGD & SGD & SGD & SGD \\
    IoU            & 0.7 & 0.65 & 0.7 & 0.7 \\
    Epoch      & 100 & 100 & 100 & 100 \\
    Batch Size            & 64 & 64 & 64 & 64 \\
    Momentum            & 0.937 & 0.0.843 & 0.937 & 0.937 \\
    Weight Decay       & 0.0005 & 0.00036 & 0.0005 & 0.0005 \\
    lr0         & 0.01 & 0.0032 & 0.01 & 0.01 \\
    lrf      & 0.01 & 0.12 & 0.1 & 0.01 \\
    Warmup Momentum            & 0.8 & 0.5 & 0.8 & 0.8 \\
    Warmup Bias lr     & 0.1 & 0.05 & 0.1 & 0.1 \\
    \hline
  \end{tabular}
\end{table}

\subsection{Category-wise Performance}
\textcolor{black}{The performance of vehicle detection using YOLO v5, v6, v7, and v8 is stated in Table VI. According to the table, YOLO v5 shows mAP0.5 is 0.826, successively YOLO v6 is 0.837, YOLO v7 is 0.841 and, YOLO v8 is 0.848. All of them YOLO v8 exhibits better performance than other models. Figure 8 (a) is the confusion matrix representation of model YOLO v8. According to the confusion matrix, class Bhotbhoti, Easybike, Leguna, Powertiller, and Wheelbarrow have much greater True Positive. In contrast, Motorbike, MPV, Pedestrian, Pickup, Truck has fewer true positives. For example, an MPV has a true positive of 0.61 which shares a large portion of false positive with a car due to its common visual appearance. As per the notation in Figure-1, the Pedestrian class has a large number of instances compared to other classes, but due to high variance in the visual appearance of Pedestrians and annotation box background selection as ground truth might be a reason that is affecting the performance. A definite proof of the claim is also represented by the native Bhotobhoti class, whose appearance is hardly matched with others. Without being a small number of representations, detection accuracy in this class is quite high. For other classes, it is either False Negative Background or False Positive Background. 
Figure 8 (b) shows the precision-recall curve of the YOLO v8 model. The precision-recall curve encapsulates the tradeoff of both metrics and maximizes the effect of both metrics, It represents a better idea of the overall accuracy of the model.
}

\begin{table}[htbp]
\resizebox{0.49 \textwidth}{!}{ 
\begin{tabular}{|c|c|c|c|c|}
\hline
Categories       & YOLOv5         & YOLOv6         & YOLOv7         & YOLOv8         \\ \hline
Bicycle          & 0.786          & 0.789          & 0.808          & 0.804          \\
Bus              & 0.899          & 0.892          & 0.912          & 0.908          \\
Bhotbhoti        & 0.926          & 0.938          & 0.944          & 0.964          \\
Car              & 0.876          & 0.881          & 0.899          & 0.881          \\
CNG              & 0.893          & 0.906          & 0.918          & 0.921          \\
Easybike         & 0.881          & 0.888          & 0.905          & 0.899          \\
Leguna           & 0.904          & 0.91           & 0.917          & 0.935          \\
Motorbike        & 0.776          & 0.795          & 0.79           & 0.796          \\
MPV              & 0.694          & 0.717          & 0.727          & 0.726          \\
Pedestrian       & 0.633          & 0.652          & 0.657          & 0.669          \\
Pickup           & 0.675          & 0.72           & 0.708          & 0.715          \\
PowerTiller      & 0.979          & 0.975          & 0.973          & 0.976          \\
Rickshaw         & 0.881          & 0.887          & 0.892          & 0.884          \\
ShoppingVan      & 0.74           & 0.755          & 0.736          & 0.763          \\
Truck            & 0.852          & 0.85           & 0.857          & 0.867          \\
Van              & 0.775          & 0.787          & 0.788          & 0.797          \\
Wheelbarrow      & 0.886          & 0.888          & 0.867          & 0.91           \\ \hline
\textbf{Overall} & \textbf{0.826} & \textbf{0.837} & \textbf{0.841} & \textbf{0.848} \\ \hline
\end{tabular}}
\caption{Category-wise Average Precision at BNVD Dataset}
\end{table}

\begin{figure}[!t]
\centering
\includegraphics[width=3.0in]{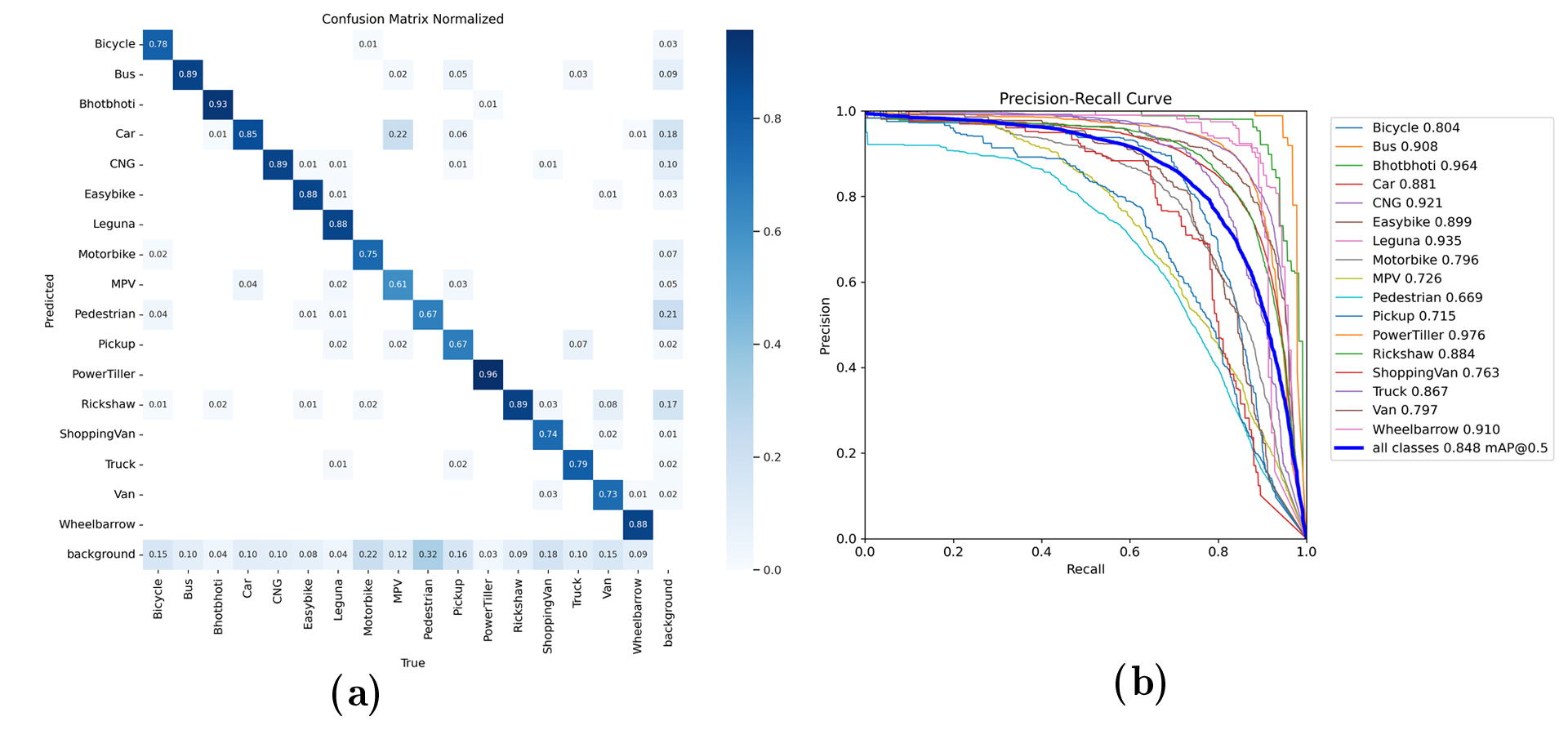}
\caption{(a) Class-wise Confusion Matrix (b) Precision-Recall Curve of YOLO v8 model}
\label{fig_8}
\end{figure}

\subsection{Dataset-wise Performance}
\textcolor{black}{Table VII presents performance evaluation metrics for all analyzed datasets. CARL-D demonstrates the highest mAP0.5 and mAP over 0.5 to 0.95 for the YOLOv6 model, with values of 0.479 and 0.372 respectively. However, when considering precision, YOLOv5 slightly outperforms. The highest recall of 0.459 for CARL-D is achieved using the YOLOv7 model. DhakaAI excels in mAP 0.5 and mAP over 0.5 to 0.95 using YOLOv7. Conversely, YOLOv8 achieves the highest precision of 0.694, while the highest recall is obtained using YOLOv6. P2 Dhaka performs best in evaluation metrics with the YOLOv6 model, achieving an mAP 0.5 of 77.5 and an mAP over 0.5 to 0.95 of 0.494. In contrast, the PoribohonBD dataset achieves its peak performance with the YOLOv5 model, obtaining an mAP of 0.981, the highest among all evaluated datasets, and reaching mAP0.5-0.95 values of 0.743, along with precision and recall values of 0.939 and 0.948 respectively.}

\begin{table}[]
\caption{Evaluation Benchmark on CARL-D, DhakaAI, P2 Dhaka, PoribohonBD, and BNVD Dataset using YOLO models.}
\resizebox{0.49 \textwidth}{!}{
\begin{tabular}{|c|c|c|c|c|c|}
\hline
Model                            & Dataset       & mAP0.5         & \begin{tabular}[c]{@{}c@{}}mAP\\ 0.5:0.95\end{tabular} & Precision      & Recall         \\ \hline
{\textbf{YOLOv5}} & CARL-D        & 0.437          & 0.328                                                  & 0.633          & 0.423          \\ \cline{2-6} 
                                 & DhakaAI       & 0.416          & 0.255                                                  & 0.640          & 0.393          \\ \cline{2-6} 
                                 & P2 Dhaka      & 0655           & 0.400                                                  & 0.804          & 0.581          \\ \cline{2-6} 
                                 & PoribohonBD   & 0.981          & 0.743                                                  & 0.939          & 0.948          \\ \cline{2-6} 
                                 & \textbf{BNVD} & \textbf{0.826} & \textbf{0.609}                                         & \textbf{0.836} & \textbf{0.762} \\ \hline
{\textbf{YOLOv6}} & CARL-D        & 0.479          & 0.372                                                  & 0.58           & 0.453          \\ \cline{2-6} 
                                 & DhakaAI       & 0.420          & 0.262                                                  & 0.311          & 0.548          \\ \cline{2-6} 
                                 & P2 Dhaka      & 0.775          & 0.494                                                  & 0.762          & 0.71           \\ \cline{2-6} 
                                 & PoribohonBD   & 0.899          & 0.648                                                  & 0.899          & 0.81           \\ \cline{2-6} 
                                 & \textbf{BNVD} & \textbf{0.837} & \textbf{0.624}                                         & \textbf{0.805} & \textbf{0.76}  \\ \hline
{\textbf{YOLOv7}} & CARL-D        & 0.478          & 0.369                                                  & 0.619          & 0.459          \\ \cline{2-6} 
                                 & DhakaAI       & 0.464          & 0.284                                                  & 0.692          & 0.438          \\ \cline{2-6} 
                                 & P2 Dhaka      & 0.743          & 0.462                                                  & 0.816          & 0.688          \\ \cline{2-6} 
                                 & PoribohonBD   & 0.907          & 0.656                                                  & 0.914          & 0.841          \\ \cline{2-6} 
                                 & \textbf{BNVD} & \textbf{0.841} & \textbf{0.623}                                         & \textbf{0.83}  & \textbf{0.779} \\ \hline
{\textbf{YOLOv8}} & CARL-D        & 0.478          & 0.359                                                  & 0.602          & 0.446          \\ \cline{2-6} 
                                 & DhakaAI       & 0.435          & 0.276                                                  & 0.694          & 0.446          \\ \cline{2-6} 
                                 & P2 Dhaka      & 0.69           & 0.449                                                  & 0.798          & 0.604          \\ \cline{2-6} 
                                 & PoribohonBD   & 0.889          & 0.658                                                  & 0.898          & 0.823          \\ \cline{2-6} 
                                 & \textbf{BNVD} & \textbf{0.848} & \textbf{0.643}                                         & \textbf{0.841} & \textbf{0.774} \\ \hline
\end{tabular}}
\end{table}

\textcolor{black}{Our proposed BNVD dataset has also been evaluated with these four YOLO models. The results showed that YOLOv8 attains the best performance, with a mean average precision of 0.847 at 0.5 IoU. YOLOv5, v6, and v7 exhibited mean average precision of 0.826, 0.837, and 0.841 respectively.}

\subsection{Lighting Condition Performance}
\textcolor{black}{The comparative evaluation of YOLOv7 and YOLOv8 models involves assessing their performance across three distinct scenarios: day, night, and limited light, as illustrated in Table VIII. YOLOv8 consistently demonstrates a slight superiority over YOLOv7 in terms of mAP0.5 across all categories. Notably, YOLOv8 excels over YOLOv7 in both day and night settings for mAP values ranging from 0.5 to 0.95. However, in limited light conditions, YOLOv7 achieves a significantly higher mAP range of 0.5 to 0.95, recording 0.942 compared to YOLOv8's 0.76. Regarding precision, YOLOv7 showcases superior performance over YOLOv8 in both daytime and nighttime scenarios. Conversely, YOLOv7 exhibits marginally better recall across all scene categories.}

\begin{table*}[htbp]
\centering
\caption{Evaluation Benchmark on Day, Limited-Light and Night scenes using best two models.}
\begin{tabular}{|c|cccc|cccc|cccc|}
\hline
{Model} & \multicolumn{4}{c|}{Day (69946 Instances)}                                                                                                     & \multicolumn{4}{c|}{Limited (5103 Instances)}                                                                                                  & \multicolumn{4}{c|}{Night (6453 Instances)}                                                                                                    \\ \cline{2-13} 
                       & \multicolumn{1}{c|}{mAP0.5} & \multicolumn{1}{c|}{\begin{tabular}[c]{@{}c@{}}mAP\\ 0.5:0.95\end{tabular}} & \multicolumn{1}{c|}{P}     & R     & \multicolumn{1}{c|}{mAP0.5} & \multicolumn{1}{c|}{\begin{tabular}[c]{@{}c@{}}mAP\\ 0.5:0.95\end{tabular}} & \multicolumn{1}{c|}{P}     & R     & \multicolumn{1}{c|}{mAP0.5} & \multicolumn{1}{c|}{\begin{tabular}[c]{@{}c@{}}mAP\\ 0.5:0.95\end{tabular}} & \multicolumn{1}{c|}{P}     & R     \\ \hline
v8                     & \multicolumn{1}{c|}{0.931}  & \multicolumn{1}{c|}{0.773}                                                  & \multicolumn{1}{c|}{0.915} & 0.869 & \multicolumn{1}{c|}{0.918}  & \multicolumn{1}{c|}{0.76}                                                   & \multicolumn{1}{c|}{0.906} & 0.849 & \multicolumn{1}{c|}{0.867}  & \multicolumn{1}{c|}{0.706}                                                  & \multicolumn{1}{c|}{0.919} & 0.781 \\ \hline
v7                     & \multicolumn{1}{c|}{0.929}  & \multicolumn{1}{c|}{0.747}                                                  & \multicolumn{1}{c|}{0.901} & 0.874 & \multicolumn{1}{c|}{0.917}  & \multicolumn{1}{c|}{0.942}                                                  & \multicolumn{1}{c|}{0.909} & 0.85  & \multicolumn{1}{c|}{0.862}  & \multicolumn{1}{c|}{0.672}                                                  & \multicolumn{1}{c|}{0.886} & 0.801 \\ \hline
\end{tabular}
\end{table*}

\subsection{Adverse Weather Condition Performance}
\textcolor{black}{Table IX displays the evaluation of the performance metrics of adverse weather condition scenes using the YOLOv8 and YOLOv7 models. Initially, we utilized pre-existing weights from a subset of the dataset, which lacked adverse weather images (Regular Images). This initial evaluation revealed subpar performance of the selected models when applied to rainy and foggy scenarios. Specifically, the YOLOv8 model yielded a mean average precision (mAP) of 0.615 at 0.5 Intersection over Union (IoU) for rainy conditions and 0.485 for foggy conditions. Similarly, YOLOv7 exhibited inferior performance with mAP0.5 values of 0.423 and 0.472 for rainy and foggy conditions, respectively.
Following the inclusion of the adverse weather subset (Regular + Adverse Images) into our dataset, notable improvements in model performance were observed. The YOLOv8 model demonstrated enhanced capabilities, achieving an mAP0.5 of 0.892 for rainy conditions and 0.922 for foggy conditions. Concurrently, YOLOv7 exhibited a similar trend, with respective mAP0.5 values of 0.86 and 0.934 for rainy and foggy conditions.
}

\begin{table*}[htbp]
\centering
\caption{Evaluation Benchmark on Rainy and Foggy scenes using best two models.}
\begin{tabular}{|c|c|cccc|cccc|}
\hline
{Model} & {Weight} & \multicolumn{4}{c|}{Rainy (5424 Instances)}                                                                                                    & \multicolumn{4}{c|}{Foggy (3396 Instances)}                                                                                                    \\ \cline{3-10} 
                       &                         & \multicolumn{1}{c|}{mAP0.5} & \multicolumn{1}{c|}{\begin{tabular}[c]{@{}c@{}}mAP\\ 0.5:0.95\end{tabular}} & \multicolumn{1}{c|}{P}     & R     & \multicolumn{1}{c|}{mAP0.5} & \multicolumn{1}{c|}{\begin{tabular}[c]{@{}c@{}}mAP\\ 0.5:0.95\end{tabular}} & \multicolumn{1}{c|}{P}     & R     \\ \hline
{v8}    & Regular                     & \multicolumn{1}{c|}{0.615}  & \multicolumn{1}{c|}{0.565}                                                  & \multicolumn{1}{c|}{0.764} & 0.56  & \multicolumn{1}{c|}{0.485}  & \multicolumn{1}{c|}{0.418}                                                  & \multicolumn{1}{c|}{0.658} & 0.439 \\ \cline{2-10} 
                       & Regular + Adverse                     & \multicolumn{1}{c|}{0.892}  & \multicolumn{1}{c|}{0.749}                                                  & \multicolumn{1}{c|}{0.899} & 0.819 & \multicolumn{1}{c|}{0.922}  & \multicolumn{1}{c|}{0.771}                                                  & \multicolumn{1}{c|}{0.931} & 0.842 \\ \hline
{v7}    & Regular                    & \multicolumn{1}{c|}{0.423}  & \multicolumn{1}{c|}{0.314}                                                  & \multicolumn{1}{c|}{0.466} & 0.427 & \multicolumn{1}{c|}{0.472}  & \multicolumn{1}{c|}{0.345}                                                  & \multicolumn{1}{c|}{0.616} & 0.471 \\ \cline{2-10} 
                       & Regular + Adverse                    & \multicolumn{1}{c|}{0.86}   & \multicolumn{1}{c|}{0.69}                                                   & \multicolumn{1}{c|}{0.889} & 0.79  & \multicolumn{1}{c|}{0.934}  & \multicolumn{1}{c|}{0.765}                                                  & \multicolumn{1}{c|}{0.904} & 0.883 \\ \hline
\end{tabular}
\end{table*}

\section{Discussion}

\textcolor{black}{The efficacy of the vehicle detection model is significantly influenced by the visual appearance of the vehicles, as demonstrated through experimental analysis. In contrast to conventional methodologies which often encounter challenges in accurately discerning object classes across diverse viewpoints, the integration of viewpoint-invariant vehicle images, alongside datasets featuring adverse weather conditions and luminance variations, confers robustness to detection even under adverse conditions, which clearly represented by Section VI (D) and (E). Furthermore, including both frontal and viewpoint-invariant roadside images helps overcome existing limitations, enhancing the variety and realism in tasks related to autonomous driving perception. By using information from different viewpoints and environmental conditions, these advancements make vehicle detection in autonomous driving systems safer and smarter, enabling them to navigate complex real-world situations accurately and reliably.}

\section{Conclusion}
The BNVD dataset covers most of the commonly appeared vehicle classes in Bangladesh and is the largest dataset in the Bangladeshi environment to our best knowledge. We established a benchmark using state-of-the-art YOLO models and meets priority requirements for being used in practical applications. The dataset contains a variety of vehicles, including some that are unique to specific regions. Additionally, some vehicles within the same class exhibit significant visual variation. \textcolor{black}{In the realm of autonomous driving datasets, the current system is constrained by its reliance solely on discrete images. However, in our forthcoming developmental endeavors, we intend to incorporate data from multiple sensors to enhance robustness within the unique environmental context of Bangladesh. Despite having limitations,} the suggested dataset presents difficulties due to its vast number of vehicle examples, varied class distributions, orientations, sizes, and complex rural and urban settings. This dataset has the potential to catalyze the development of machine learning models for object detection in computer vision, thereby addressing current challenges.



%


\begin{IEEEbiography}[{\includegraphics[width=1in,height=1.25in,clip,keepaspectratio]{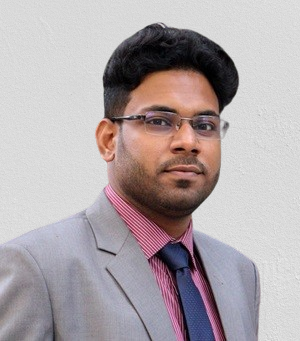}}]{Bipin Saha}
obtained his B.Sc. in Electrical and Electronic Engineering from the University of Rajshahi, Bangladesh, in 2022. Presently, he is working as a Machine Learning Engineer at Business Automation Ltd., Rajshahi. His research interests revolve around Computer Vision, Robot Motion Planning, Autonomous Navigation, Visual Perception, and the development of Intelligent Systems. \end{IEEEbiography}
\begin{IEEEbiography}[{\includegraphics[width=1in,height=1.25in,clip,keepaspectratio]{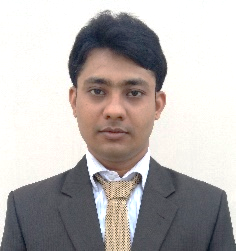}}]{Md. Johirul Islam}
was born in Meherpur, Bangladesh in 1989. He received his Ph.D. degree in 2022 from the Faculty of Engineering, University of Rajshahi, Bangladesh. He received his B.Sc. and M.Sc. degrees in Applied Physics and Electronic Engineering from the same university in 2011 and 2012, respectively. Currently, he is an Assistant Professor in the Department of Physics, Rajshahi University of Engineering and Technology, Bangladesh. His research interests include feature engineering, EMG pattern recognition, human machine interfacing, biomedical instrumentation, and embedded systems. \end{IEEEbiography}
\begin{IEEEbiography}[{\includegraphics[width=1in,height=1.25in,clip,keepaspectratio]{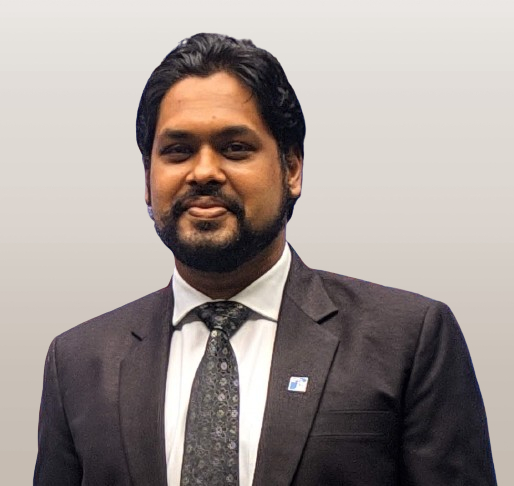}}]{Shaikh Khaled Mostaque}
Skhaikh Khaled Mostaque received the B.Sc. and M.Sc. degrees in Applied Physics and Electronic Engineering from the University of Rajshahi, Rajshahi,
Bangladesh. He is currently working an Assistant Professor with the Department
of Electrical and Electronic Engineering, University of Rajshahi, Rajshahi. He is also serving as Advisor at IEEE Robotics and Automation Society University of Rajshahi Student Branch Chapter. His research interests include computer vision, embedded systems, renewable energy materials and photovoltaics.\end{IEEEbiography}
\begin{IEEEbiography}[{\includegraphics[width=1in,height=1.25in,clip,keepaspectratio]{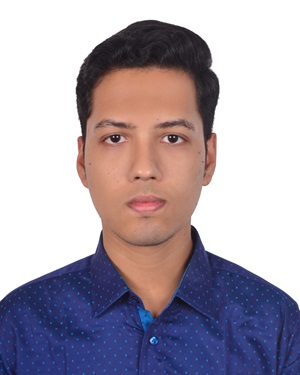}}]{Aditya Bhowmik}
is an undergraduate student pursuing Electrical and Electronic Engineering at the University of Rajshahi. As an active member of the IEEE and serving as a secretary at the University of Rajshahi Student Branch Chapter of the IEEE Robotics and Automation Society, he's deeply engaged in the field. Aditya's passion for research is evident, focusing on robotics, automation, and the advancements in solar cell technology.\end{IEEEbiography}
\begin{IEEEbiography}[{\includegraphics[width=1in,height=1.25in,clip,keepaspectratio]{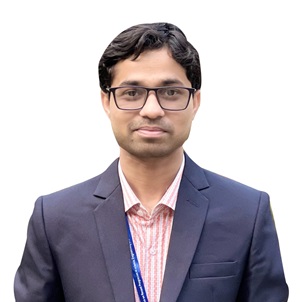}}]{Tapodhir Karmakar Taton} 
is pursuing an undergraduate degree in Electrical and Electronic Engineering at the University of Rajshahi. He is currently serving as a Chairperson at IEEE Robotics and Automation Society University of Rajshahi Student Branch Chapter. His primary research interests include Robotics, Machine Learning, Computer Vision, IoT, and Embedded Systems.\end{IEEEbiography}
\begin{IEEEbiography}[{\includegraphics[width=1in,height=1.25in,clip,keepaspectratio]{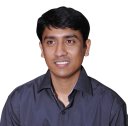}}]{Md Nakib Hayat Chowdhury}
 a distinguished individual in the field of Computer Science and Engineering, completed his BSc. and MSc. degrees at the University of Rajshahi, establishing a solid foundation in his academic journey. From 2012 to 2015, he served as a Lecturer at Varendra University, Rajshahi, where his passion for computer science education became evident. During this period, he played a crucial role in shaping the academic experiences of aspiring students. In 2015, Mr. Chowdhury transitioned to the Bangladesh Army University of Science and Technology, continuing his role as a dedicated Lecturer until 2018. His commitment to excellence in education and his expertise in computer science made a lasting impact on the university community. Recognizing his contributions, he was elevated to the position of Assistant Professor at the Bangladesh Army University of Science and Technology in 2018. Throughout the subsequent five years, until 2023, he continued to inspire and educate students, leaving an indelible mark on the institution. Md Nakib Hayat Chowdhury's journey is a testament to his commitment to advancing education in Computer Science and Engineering, and his legacy resonates through the students whose academic lives he enriched during his tenure. Currently he is pursuing his PhD in Biomedical Engineering at the National University of Malaysia.\end{IEEEbiography}
\begin{IEEEbiography}[{\includegraphics[width=1in,height=1.25in,clip,keepaspectratio]{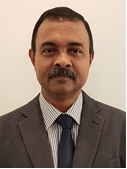}}]{
Mamun Bin Ibne Reaz}
 PhD ’07, is the Dean of the School of Engineering, Technology and Sciences at the Independent University, Bangladesh, and Professor in Electrical and Electronic Engineering. Previously, he was a Professor at the Universiti Kebangsaan Malaysia, Malaysia. His scientific specialization is in the areas of IC Design, Biomedical application IC, Biomedical sensors, and Smart Home. Mamun Bin Ibne Reaz has published more than 400 scientific articles and is a recipient of more than 70 research grants. Since 2020, he is listed among the world's top 2\% of scientists by Stanford University Data for "Updated science-wide author databases of standardized citation indicators". He was a Senior Associate of the Abdus Salam International Centre for Theoretical Physics (ICTP), Italy since 2008, and is presently, the Coordinator of the ICTP EAU Affiliated Centre in UKM, Malaysia. Mamun Bin Ibne Reaz has an undergraduate and graduate degree in Applied Physics and Electronics from the University of Rajshahi, Bangladesh, and a doctoral degree in VLSI Design from the Ibaraki University, Japan.\end{IEEEbiography}

\vspace{11pt}

\vfill

\end{document}